\pdfoutput=1

\documentclass[11pt]{article}

\usepackage{ACL2023}

\usepackage{times}
\usepackage{latexsym}
\usepackage{multirow}
\usepackage{tabularx}
\usepackage{colortbl}
\usepackage[T1]{fontenc}

\usepackage[utf8]{inputenc}

\usepackage{microtype}
\usepackage{graphicx}
\usepackage{booktabs}
\usepackage{color}
\usepackage{makecell}
\usepackage{diagbox}
\usepackage{booktabs}
\usepackage{setspace}

\usepackage{inconsolata}
\usepackage{xcolor}

\usepackage{array}
    \newcolumntype{P}[1]{>{\arraybackslash}p{#1}}
    \newcolumntype{M}[1]{>{\centering\arraybackslash}m{#1}}

\usepackage[framemethod=TikZ]{mdframed}
\newenvironment{RecBox}[1]
  {\mdfsetup{
    frametitle={\colorbox{white}{\space#1\space}},
    innerleftmargin = 0.25cm, innerrightmargin = 0.25cm, innertopmargin = 0cm, innerbottommargin = 0.25cm,
    frametitleaboveskip=-\ht\strutbox,
    frametitlealignment={\hspace{-5pt}},
    skipabove=10pt,
    skipbelow=7pt
    }
  \begin{mdframed}%
  
  }
  {\end{mdframed}}

\definecolor{lightgreen}{RGB}{144,238,144}
\definecolor{lightred}{RGB}{255,187,187}
\definecolor{lightorange}{RGB}{255,223,155}
\definecolor{lightblue}{RGB}{173, 216, 230}

\title{From Languages to Geographies: Towards Evaluating \\ Cultural Bias in Hate Speech Datasets
}

\author{Manuel Tonneau\textsuperscript{\rm 1, \rm 2, \rm 3}, 
        Diyi Liu\textsuperscript{\rm 1},
        Samuel Fraiberger\textsuperscript{\rm 2, \rm 3, \rm 4}, \\
        {\bf Ralph Schroeder}\textsuperscript{\rm 1},
        {\bf Scott A. Hale}\textsuperscript{\rm 1,\rm 5}, 
        {\bf Paul Röttger}\textsuperscript{\rm 6}\\
        \textsuperscript{\rm 1}University of Oxford,
        \textsuperscript{\rm 2}World Bank,
        \textsuperscript{\rm 3}New York University, \\
        \textsuperscript{\rm 4}Massachusetts Institute of Technology,
        \textsuperscript{\rm 5}Meedan,   
        \textsuperscript{\rm 6}Bocconi University
}

\begin{document}
\maketitle
\begin{abstract}
Perceptions of hate can vary greatly across cultural contexts. Hate speech (HS) datasets, however, have traditionally been developed by language. This hides potential cultural biases, as one language may be spoken in different countries home to different cultures. In this work, we evaluate cultural bias in HS datasets by leveraging two interrelated cultural proxies: language and geography. We conduct a systematic survey of HS datasets in eight languages and confirm past findings on their English-language bias, but also show that this bias has been steadily decreasing in the past few years. For three geographically-widespread languages---English, Arabic and Spanish---we then leverage geographical metadata from tweets to approximate geo-cultural contexts by pairing language and country information. We find that HS datasets for these languages exhibit a strong geo-cultural bias, largely overrepresenting a handful of countries (e.g., US and UK for English) relative to their prominence in both the broader social media population and the general population speaking these languages. Based on these findings, we formulate recommendations for the creation of future HS datasets. 

\end{abstract}

\section{Introduction}

Far from the idyllic image of social media connecting people, increasing social cohesion, or letting everyone have an equal say, harmful content including hate speech (HS) has become rampant online \cite{vidgen2019much} and has been linked to social unrest, hate crimes, and even deaths \citep{banaji2019whatsapp,muller2021fanning}. 

\begin{figure}[t]
    \raggedleft
    \includegraphics[width=0.5\textwidth]{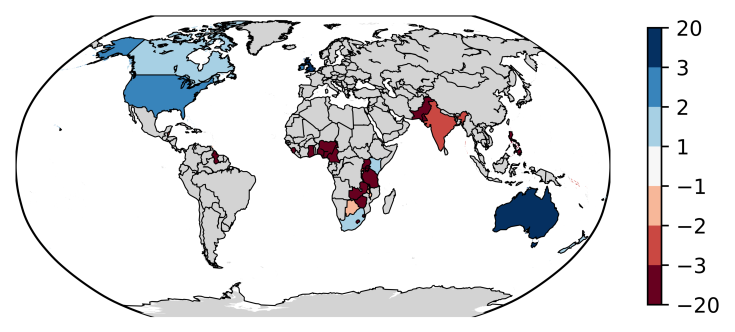} 
    \vspace*{-2em}
    \caption{Geographical representativeness of author population of English hate speech datasets. A \colorbox{lightblue}{positive value} $N$ (\colorbox{lightred}{negative value} $-N$) indicates that a country is $N$ times \colorbox{lightblue}{more} (\colorbox{lightred}{less}) represented in English hate speech datasets relative to the global English-speaking population. }
    \vspace*{-1em}

    \label{fig:first_page}
\end{figure} 

To counter this phenomenon, a mature body of research has developed annotated datasets for automatic HS detection \cite{vidgen2020directions}. Past work, however, has highlighted systematic gaps and biases in HS datasets \cite{park-etal-2018-reducing,davidson-etal-2019-racial, wiegand-etal-2019-detection, nejadgholi-kiritchenko-2020-cross, wich-etal-2020-impact}. In particular, HS datasets exhibit a strong language bias, with the vast majority of datasets developed for English \cite{poletto2021resources}. This focus on English, and more generally on languages, when developing HS datasets creates a risk of cultural blindness. Indeed, while certain languages, such as Basque, Icelandic or Yoruba, are highly indicative of a certain cultural context, others, such as English, are present across cultures. Yet, understanding the cultural context of a statement is crucial to determine whether it is hateful \cite{aroyo2019crowdsourcing}. Statements may be perceived as hateful in one culture but not in another \cite{lee-etal-2023-hate}, even within the same language \cite{lee2023crehate}. For instance, the term ``Paki'' is used as a neutral abbreviation for Pakistani in Pakistan whereas it is a racial slur in the UK. Despite the importance of the cultural context in the study of HS, the cultural origin of HS datasets remains largely unclear.

In this work, we aim to bridge this gap by answering the following research question: \textbf{To what extent are HS datasets culturally biased?} We operationalize cultural bias by measuring the representation of two cultural proxies in HS datasets: (a) language, and (b) geo-cultural contexts \cite{de2018evaluative}, defined as the combination of a language and a country. We first conduct a systematic survey of HS datasets in eight widely-spoken languages: Arabic, English, French, German, Indonesian, Portuguese, Spanish and Turkish. We confirm past findings on their English-language bias but also show that this dominance has been steadily decreasing in the past few years, with other languages such as Arabic catching up. We then depart from the traditional language-level analysis and situate our analysis in geo-cultural contexts. We focus on three geographically-widespread languages---English, Arabic and Spanish---and on Twitter, the main data source for HS datasets. We leverage geographical metadata from the annotated tweets in the datasets to infer the locations of their authors and find that HS datasets for these languages predominantly represent authors from a handful of countries (the US and UK for English, Chile and Spain for Spanish, and Jordan for Arabic). We also find that such countries are largely overrepresented in HS datasets compared to their prominence in both the broader social media population and the general population speaking these languages. We identify two main factors to explain the lack of representativeness of HS datasets: the lack of representativeness of Twitter itself as well as the sampling decisions made by authors. For the latter, we observe that non-uniform geographic sampling is typically intentional for Arabic and Spanish, motivated by a focus on specific geo-cultural contexts. In contrast, we find that such non-uniform sampling is commonly disregarded when compiling English HS datasets, which systematically lack information on the geographical origin of both data and annotators, hiding potential mismatches and ignoring the diversity of English speakers online. Based on these findings, we formulate recommendations for the creation of future HS datasets. Overall, our main contributions are:

\begin{enumerate}
\itemsep=0em
\vspace*{-0.5em}

\item A systematic survey of 75 HS datasets in eight languages (Arabic, English, French, German, Indonesian, Portuguese, Spanish and Turkish), revealing a persistent, though diminishing, dominance of English (\S\ref{sec:language_bias}).

\item Evidence of geo-cultural bias in existing HS datasets for three geographically-widespread languages: English, Arabic and Spanish (\S\ref{sec:geo_cultural_bias}).
\item Preprocessed HS corpora for the eight surveyed languages and code for geocoding to stimulate research in this area.\footnote{\url{https://github.com/manueltonneau/hs-survey-cultural-bias}} 
    
\end{enumerate}

\section{Background}

\subsection{Languages and Geographies as Interrelated Cultural Proxies}

Language has historically played a pivotal role in cultural identity \cite{collins1999macrohistory} and can be a good proxy for culture when a certain language is spoken only by a specific cultural group (e.g., Basque).  Yet, some languages, such as English, Arabic or Spanish, have transcended cultural boundaries through human mobility, colonization, and imperialism. Such global adoption means that people who share a common language may come from diverse cultural backgrounds. These cultural differences also have online implications, whereby social media communities tend to form around both a common language and geography rather than just a common language \cite{mekacher2024language}. To take into account such differences, we use both language and geo-cultural contexts in our analysis of cultural bias. Cross-language bias measures how well different languages are represented, while geo-cultural contexts capture the representation of geographic
locations, taking into account the cultural characteristics of a population, such as a common language \cite{de2018evaluative}.

\subsection{Cultural Biases in NLP}

The drastic progress in NLP tasks over the past decade can be partially attributed to the growing availability of large text corpora \cite{raffel2020exploring}, used to train language models. Yet, past work shows that these corpora are largely composed of English-language content \cite{joshi-etal-2020-state, holtermann2024evaluating, zhao2024large}, containing smaller amounts and lower-quality content for other widely spoken languages \cite{kreutzer-etal-2022-quality}. Adding to such language biases, past work has uncovered geographic biases in NLP corpora, where represented dialects and topics disproportionately originate from the Minority World \cite{graham2014uneven,graham2015digital,dodge-etal-2021-documenting}. Driven by the necessity to include social factors in language modeling \cite{hovy-yang-2021-importance}, an emerging body of scholarship has developed approaches to include geographical information in language representation \cite{bamman-etal-2014-distributed, rahimi-etal-2017-continuous, hovy-purschke-2018-capturing, kulkarni-etal-2021-lmsoc-approach, hofmann2022geographic}. Despite these efforts, recent language models still suffer from cultural biases, mirroring views largely aligned with Western, Educated, Industrialized, Rich and Democratic (WEIRD) individuals \cite{atari2023humans,naous2023having,manvi2024large}. In order to mitigate such biases, it is crucial to document their presence in training and evaluation corpora, especially for culturally-sensitive tasks like HS detection \cite{baider2020pragmatics}.

\subsection{Biases in Hate Speech Datasets}

Past work has highlighted several biases in HS datasets. Many such biases can be linked to the subjectivity and demographics of annotators \cite{al-kuwatly-etal-2020-identifying}, including racial bias  \cite{davidson-etal-2019-racial, sap-etal-2019-risk}, gender bias \cite{park-etal-2018-reducing}, and political bias \cite{wich-etal-2020-impact}. Other biases are related to the way such datasets are constructed, resulting in a large overrepresentation of the hateful class as well as certain topics and users \cite{dixon2018measuring,davidson-etal-2019-racial, wiegand-etal-2019-detection,nejadgholi-kiritchenko-2020-cross}. Despite the extent of this scholarship, little attention has been given to cultural bias in HS corpora. The most recent widely-cited and large-scale survey of HS resources does point to an English-language bias \cite{poletto2021resources} and a dominance of Twitter as a data source, which is known to be skewed towards certain geo-cultural contexts.\footnote{\url{https://datareportal.com/essential-twitter-stats}} Also, \citet{arango-monnar-etal-2022-resources} point out that Spanish HS datasets are largely developed in the national context of Spain, motivating tailored approaches to other Spanish-speaking contexts such as Chile. Finally, past work highlights the cultural sensitivity of HS, uncovering country-specific offensive words \cite{ghosh-etal-2021-detecting} as well as disparities in cross-cultural HS annotations \cite{lee2023crehate}, stereotype definition \cite{bhutani2024seegull} and cross-dialect HS detection performance \cite{castillo-lopez-etal-2023-analyzing} for a given language. To the best of our knowledge, our work is the first to systematically investigate cultural bias in HS datasets.

\section{Language Bias in Hate Speech Datasets}\label{sec:language_bias}




\begin{table}[htp]
\centering
\footnotesize
\resizebox{0.48\textwidth}{!}{%
\begin{tabular}{lrrrrr}
\toprule
 \textbf{Language} & \makecell{Twitter\\only} & \makecell{Twitter +\\other} & \makecell{Other} & \makecell{Synthetic} & \textbf{Total}  \\
\midrule 
English    & 12 & 3 & 10 & 4 &  \textbf{29} \\
Arabic & 11 & 0 & 0 & 1 &  \textbf{12} \\
Spanish   & 6 & 0 & 0 & 1 &  \textbf{7} \\
German   & 2 & 1 & 2 & 2 &  \textbf{7} \\
Turkish & 5 & 0 & 1 & 0 &  \textbf{6}\\
French   & 3 & 0 & 1 & 2 & \textbf{6} \\
Portuguese   & 3 & 0 & 1 & 1 &  \textbf{5} \\
Indonesian   & 2 & 0 & 1 & 0 & \textbf{3} \\
\bottomrule
\end{tabular}}
\caption{Number of available hate speech datasets by language and data source}
\vspace*{-1em}
\label{tab:dataset_survey}
\end{table}

We start our analysis of cultural bias at the language-level, as some languages are specific to single cultural contexts. We conduct a systematic survey of HS datasets in eight languages with a large presence on social media platforms: Arabic, English, French, German, Indonesian, Portuguese, Spanish and Turkish.

\subsection{Survey Approach}
To identify HS datasets, we rely on three data sources. First, we inspect the Hate Speech Data Catalogue\footnote{\url{https://hatespeechdata.com/}} \cite{vidgen2020directions} and find 80 candidate datasets for our languages of interest. 
Second, we inspect the datasets listed in the latest survey of HS datasets \cite{poletto2021resources} and find 20 additional candidate datasets that are not listed in the HS Data Catalogue. 
Finally, we conduct a Google search for each language and inspect the links of the first three result pages in each case, adding 43 candidate datasets that are neither in the HS Data Catalogue nor listed by \citet{poletto2021resources}.
From those 143 unique datasets, we keep only the datasets that fit the following three criteria: 
 \begin{enumerate}
     \vspace*{-0.5em}
  \itemsep-0em 
     \item The dataset is documented, meaning it is attached to a research paper or a README file describing its construction.
     \item The dataset is either publicly available or could be retrieved after contacting the authors.
     \item The dataset focuses on HS, defined broadly as ``any kind of communication in speech, writing or behavior, that attacks or uses pejorative or discriminatory language with reference to a person or a group on the basis of who they are, in other words, based on their religion, ethnicity, nationality, race, color, descent, gender or other identity factor'' \cite{un2019}.
    \vspace*{-0.5em}

 \end{enumerate} 
 We provide additional details on the surveying in the Appendix (\S \ref{sec:appendix_data_survey}). 

\subsection{Results}
Out of the 143 aforementioned datasets, we identify 75 available datasets that meet our three criteria for the eight languages of interest. We provide a breakdown in terms of language and data source in Table \ref{tab:dataset_survey} as well as the number of datapoints by language (Table \ref{tab:post_count}) and a complete list of the datasets for each language (\S\ref{sec:appendix_datasets}) in the Appendix.

\paragraph{Language and data source} We find that English is the most common language in terms of HS detection resources, representing 39\% of all available corpora and 41\% of all annotated datapoints for our eight languages of interest. We also find that Twitter is by far the most common data source across languages. This is particularly the case for Arabic, with 92\% of corpora originating from Twitter, followed by Spanish (86\%) and Turkish (83\%). Additionally, we find that some languages are particularly affected by a lack of data availability. For instance, 50\% of identified Indonesian datasets and 38\% of identified Portuguese datasets could not be retrieved (see Appendix \S\ref{sec:appendix_unavail_datasets} for more details).

\begin{figure}
    \centering
    \includegraphics[width=0.41\textwidth]{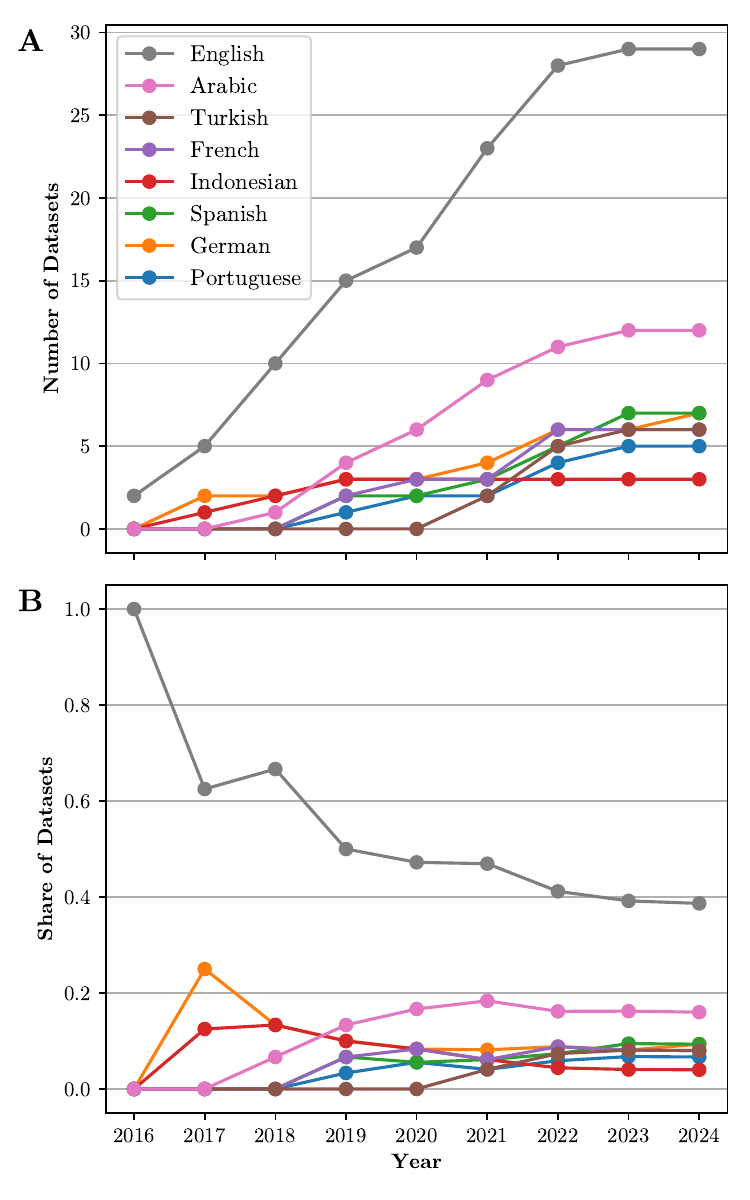} 
    \caption{(A) Number of hate speech datasets per language over time (B) Share of hate speech datasets for the 8 languages of interest over time}
    \vspace*{-1.5em}

    \label{fig:time_analysis}
\end{figure}

\vspace*{-0.5em}

\paragraph{Temporal dynamics} To understand the dynamics of HS detection resource creation across languages, we further present the number of datasets per language over time as well as the language-level share of all datasets over time (Figure \ref{fig:time_analysis}). We find that while English has dominated other languages in terms of the number of datasets over time, its dominance in terms of share of all HS datasets has steadily declined over the years, going from 100\% of all datasets for the eight languages of interest in 2016 to 39\% in 2023. In parallel, languages such as Arabic have been catching up. Such growth in corpus availability points towards a broadening of research that aims to address the multilingual nature of HS.

\section{Geo-Cultural Bias in Hate Speech Datasets}\label{sec:geo_cultural_bias}

While such language-level analysis is crucial to uncover gaps in existing resources and motivate the development of resources for under-served languages, it cannot account for and may hide potential large differences in resources between countries with a common language. In this section, we investigate the extent of geo-cultural bias in HS datasets, approximating geo-cultural contexts as a combination of one language and one country. For this purpose, we leverage the rich geographical metadata of tweets to map posts and their authors to a country location.  We focus on three geographically widespread languages---English, Arabic and Spanish---for which the HS detection resources mostly emanate from Twitter (Table \ref{tab:dataset_survey}).

\subsection{Author Location Inference}
\label{subsec:author_location_inference}

We use tweet geographical metadata to infer the country location of tweets' authors. 
 \vspace*{-0.5em}

\paragraph{Information sources} While there is a plethora of available information to infer user location from, from self-reported location to geocoordinates, timezone and linguistic features of tweets, each of these features has weaknesses. Profile locations are only available for a fraction of users, may contain vague locations (e.g., Planet Earth) or non-geographic text \cite{hecht2011tweets} and may not always match with the device location \cite{graham2014world}. Geo-coordinates are even rarer (1--2\% of all tweets according to Twitter\footnote{\url{https://developer.twitter.com/en/docs/tutorials/advanced-filtering-for-geo-data}}) and may point to locations other than a user's home location, for instance if the user is travelling. Further, linguistic features have proven to not be a good proxy for location \cite{graham2014world} and while dialectal variability may inform on a user's location \cite{jurgens-etal-2017-incorporating}, language identification methods incorporating this variability are scarce beyond English. Finally, timezones of different countries with a common language may overlap. While acknowledging these limitations, we decide to use exclusively the two features that are equally available across languages to infer user country location: the geocoordinates of tweets and the self-reported user profile location.

\begin{table}[h]
\centering
\footnotesize
\resizebox{\linewidth}{!}{
\renewcommand*{\arraystretch}{1.3}
\begin{tabular}{>{\centering\arraybackslash} m{4cm}m{1cm}m{1cm}m{1cm}}
\toprule
 & \textbf{English} & \textbf{Arabic} & \textbf{Spanish} \\ \midrule
Share of all Twitter datasets with retrieved tweet IDs & 9/15 & 6/11 & 4/6 \\
\# unique tweets with tweet IDs & 155,974 & 456,892 & 24,752  \\
\# tweets with tweet IDs and retrieved geographical metadata & 64,057 &  251,178 & 14,684  \\
\# tweets with inferred author country location      & 50,116 & 247,408 & 13,273  \\ \bottomrule
\end{tabular}}%
\caption{Summary statistics of data collection and author location inference}
\label{tab:summary_statistics}
\end{table}

\vspace*{-0.5em}

\paragraph{Geographical data collection} Tweet geocoordinates and user profile location are usually not shared in public HS datasets for privacy reasons. In this context, we first attempt to retrieve the tweet IDs of all Twitter datasets for English, Arabic and Spanish by either collecting them when they are publicly available or contacting the authors to request access. We are able to retrieve tweet IDs for 9 English \cite{waseem-2016-racist, waseem-hovy-2016-hateful, jha-mamidi-2017-compliment, elsherief2018hate, elsherief2018peer, vidgen-etal-2020-detecting, mathew2021hatexplain, samory2021call, toraman-etal-2022-large},  6 Arabic \cite{albadi2018they, alsafari2020hate, alshaalan-al-khalifa-2020-hate, mulki2021working, ameur2021aracovid19, ahmad2023hate} and 4 Spanish \cite{pereira2019detecting, garcia2021detecting, arango-monnar-etal-2022-resources, vasquez-etal-2023-homo} Twitter HS datasets. We then use the Twitter API to retrieve the tweet author self-reported location and the tweet geocoordinates if available. Out of all tweet IDs, we are able to retrieve some geographical information, that is either the tweet's author self-reported location, geocoordinates or both, for 64,057 (41\%) English, 251,178 (55\%) Arabic and 14,684 (59\%) Spanish tweets. We report the main statistics of data collection in Table \ref{tab:summary_statistics}.

\paragraph{Country inference} We infer the country of origin of a tweet author in two ways.
First, in case a tweet is geotagged, we assign the country location of the geotag to its author. In cases where a user has no geotagged tweets but has a self-reported location, we use geocoding to convert the reported location to a country location. Specifically, we use the Google Geocoding API as \citet{graham2014world} demonstrate it performs better than other geocoding tools. In case a tweet has no available geographical metadata, we are not able to infer its author country location and do not analyse it further. 

\paragraph{Geocoding evaluation} 

For each language, we sample 50 unique user locations geocoded within a country and have one author annotate whether this country match is correct. We also sample 50 unique user locations that could not be associated with a country and annotate whether they could have been associated from the information they contained. We find that the Google Geocoding API is able to associate approximately two thirds of unique user locations to a country, a value that is relatively constant across languages. We also find that this geocoding method exhibits a very high precision (92--96\% across languages), with the few errors happening for ambiguous location strings containing multiple locations and which are therefore not geocodable. Also, the share of non-geocoded user locations that could have been geocoded from the provided information is relatively low (12--16\%). These instances typically involve the use of emojis, such as national flags, and nicknames for locations (e.g., ``Down Under'' for Australia),  which the Geocoding API fails to recognize. We provide more information on the geocoding evaluation in the Appendix (\S\ref{appendix_geocoding_eval}). 

\paragraph{Inference} In total, we are able to infer the country location of 50,116 English tweets, representing 8\% of all posts from the surveyed English HS datasets, 247,408 Arabic tweets (52\%) and 13,273 Spanish tweets (27\%).

\begin{figure*}[ht]
    \raggedleft
    \includegraphics[width=1\textwidth]{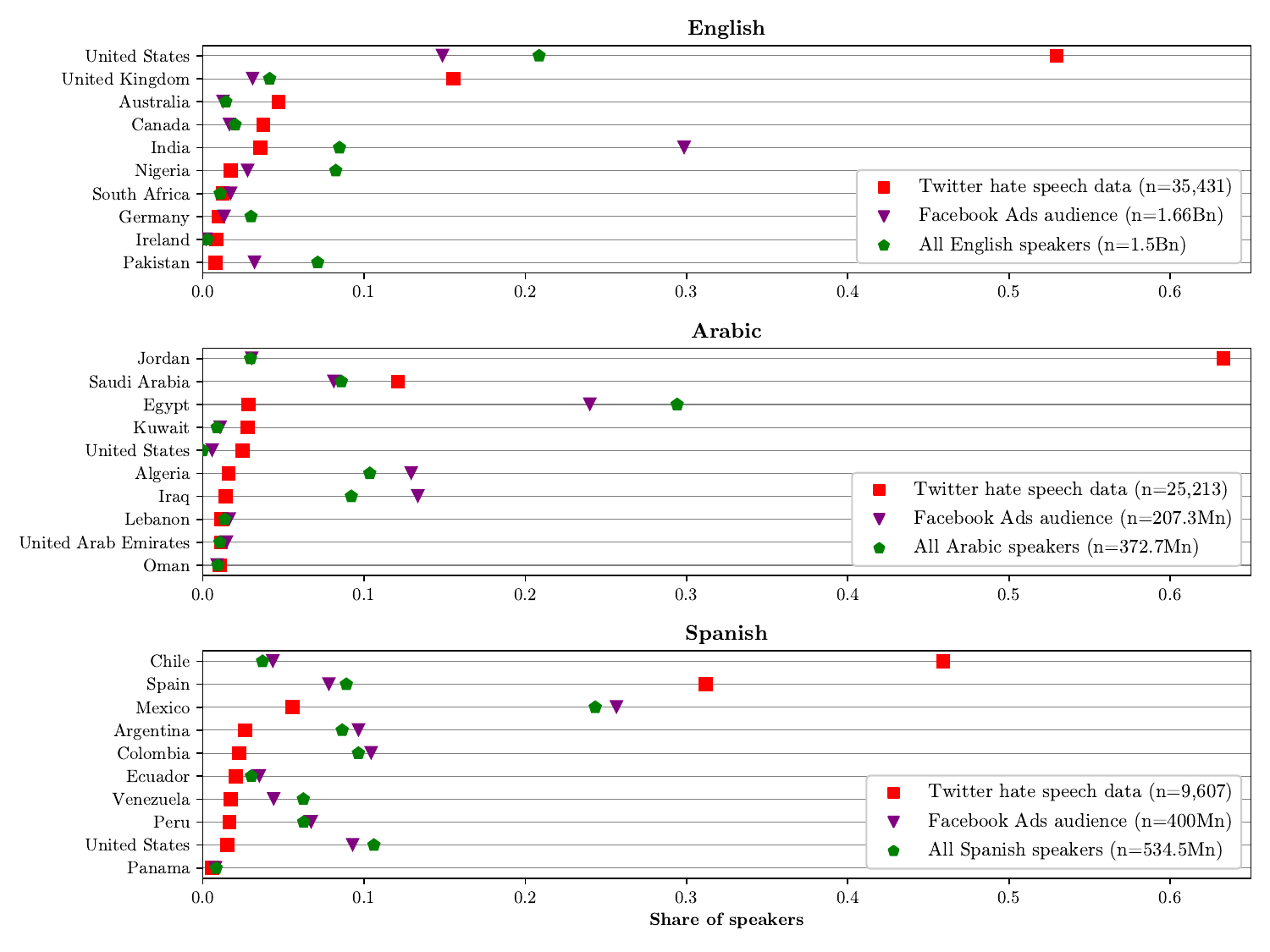} 
    \caption{Share of speakers by country location in three reference populations: Twitter users who authored the posts in the Twitter public hate speech datasets (Twitter hate speech data); Facebook and Instagram users (Facebook Ads audience) and all speakers of a language (All [language] speakers). }
    \label{fig:masterplot_cross_country}
\end{figure*}

\subsection{Reference Points for Representativeness} 

For each language $L$, we aim to assess the geo-cultural representativeness of Twitter HS datasets relative to three larger groups: the general Twitter user population speaking language $L$, the general social media population speaking $L$, and the general population of speakers of $L$. 

\paragraph{Twitter user population} In the absence of reliable information on country share of Twitter users by language, we derive this statistic by using a large Twitter dataset stemming from a recent collaborative project \citep{pfeffer2023just} that collected all tweets posted within a 24-hour period starting on September 21, 2022, including the geographical metadata. This so-called \textbf{Twitter Day} dataset amounts to approximately 116 million English tweets, 27 million Spanish tweets and 19 million Arabic tweets posted by 17, 5 and 2 million users respectively. 

\paragraph{General social media population} Given their large user population and geographical coverage,\footnote{\url{https://datareportal.com/social-media-users}} we use the Facebook and Instagram user populations as a proxy for the general social media population. Specifically, we use the audience measurement tool of \textbf{Facebook Ads}. This tool, which has been used in past demographic research \cite{zagheni2017leveraging,palotti2020monitoring, rama2020facebook}, provides the number of Facebook and Instagram users in a given country aged 13 and older that are using these platforms in each of our languages of interest. We then compute the country-level share of the overall Facebook Ads audience for each language. 

\paragraph{General population} Finally, we use official statistics on the country-level number of speakers of each language of interest. We provide further details on the data sources for each language in the Appendix (\S\ref{sec:appendix_official_stats}).

\subsection{Results}

We compute the country share of users speaking each language of interest from four different populations: (i) the Twitter users who authored the posts of the public Twitter HS datasets, (ii) the Twitter user population from the Twitter Day dataset, (iii) the broader social media population using Facebook and Instagram user populations as a proxy, and (iv) the full population of speakers of the language of interest. We report the comparison between (i), (iii), and (iv) in Figure \ref{fig:masterplot_cross_country} and between (i) and (ii) in Figure \ref{fig:masterplot_cross_country_twitter_day} in the Appendix.

\vspace*{-0.5em}

\paragraph{Bias and lack of representativeness} We observe that the majority of Twitter users who authored the posts from the HS datasets originate from a handful of countries for each language, namely the United States and the United Kingdom for English, Jordan for Arabic, and Chile and Spain for Spanish. We also find that the Twitter user population who authored the posts from the public HS datasets is a highly skewed subset of both the broader social media population and the general speaker population in terms of country location. We further observe a general trend where countries with higher economic development are overrepresented in HS datasets compared to both the social media population and the general population of speakers (notably the US, UK, Australia, and Canada for English, Spain and Chile for Spanish and to a lesser extent, Saudi Arabia and Kuwait for Arabic). In contrast, countries with lower economic development tend to be under-represented in the HS datasets (e.g., India, Nigeria, and Pakistan for English, Egypt, Algeria and Iraq for Arabic and Colombia, Venezuela and Peru for Spanish).

\paragraph{Factors affecting representation} Several factors could explain such lack of representativeness. First, the country representation in the Twitter HS data generally aligns with the country representation in the general Twitter population, which is also not representative of the broader social media population nor the total population of speakers. This is particularly the case for English (Pearson correlation of 0.99) but less the case for Spanish (0.43) and Arabic (0.21). Second, this misalignment can also be explained by sampling decisions made when creating the HS datasets. We observe that these decisions are largely intentional for Arabic and Spanish, motivated by the focus on a specific geo-cultural context. For instance, Jordan's dominance for Arabic is largely explained by the focus on users with a location in Jordan in the sampling of the largest Arabic HS dataset \cite{ahmad2023hate}. Similarly, the importance of Chile for Spanish is driven by the choice of Chilean Spanish keywords used for sampling in \citet{arango-monnar-etal-2022-resources}. 
In the case of English, sampling also appears to affect representation as we observe large gaps between the country representation in the HS datasets and in the general Twitter population (Figure \ref{fig:masterplot_cross_country_twitter_day}). Yet, such decisions appear to be either implicit or unintentional as a country focus is almost never mentioned in English HS datasets.

\begin{figure}
    \raggedleft
    \includegraphics[width=0.5\textwidth]{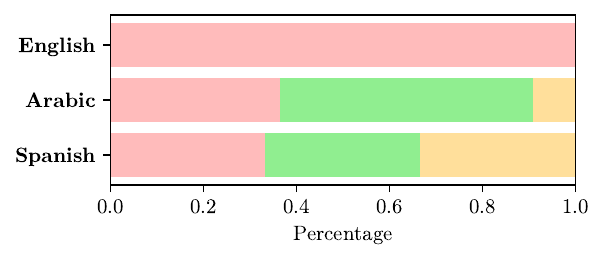} 
    \caption{Percentage share (\%) of each scenario when comparing data and annotator geographical origin: \colorbox{lightred}{no information} if either the origin of the data or of the annotator is not provided; \colorbox{lightorange}{partial alignment} if data and annotator origin partly overlap (e.g., Spanish annotators annotate tweets from Spain and Mexico) and \colorbox{lightgreen}{full alignment} if data and annotator origin perfectly overlap.}
    \label{fig:annotator_data_origin}
\end{figure}

\paragraph{Data and annotator origin} Cultural misalignment between data and annotator origin creates a risk of annotation error, due to a lack of cultural understanding. Using the information provided by the dataset authors, we measure the alignment between data and annotator origin for all non-synthetic English, Arabic and Spanish datasets. We report the results in Figure \ref{fig:annotator_data_origin}.

Our most striking result is the lack of information provided by English HS dataset creators about potential cultural misalignment. Indeed, whereas both the data and annotator origin are provided and partially or fully align in 66\% of cases for Spanish and 63\% for Arabic, none of the surveyed English datasets provide both pieces of information. Specifically, the vast majority of English HS datasets report only the data source (e.g., Twitter) but no precise geographical origin. Similarly, annotator origin is provided in most cases but usually only contains the name of the crowdsourcing platform used (e.g., MTurk, Crowdflower), whose workers originate from a variety of geographies \cite{difallah2018demographics}. 

\section{Discussion and Recommendations}

\paragraph{Bias evaluation} 
In this work, we evaluated cultural bias in HS datasets in two steps: at the language level and at the geo-cultural level, approximated as a combination of one language and one country. 
At the language-level, we observe a dominance of English in the number of HS datasets but find that this dominance has been decreasing, with other languages such as Arabic catching up. We also observe that the vast majority of HS corpora originate from Twitter. This is in line and complements the most recent widely-cited and large-scale survey of HS resources \cite{poletto2021resources}. Focusing on three geographically widespread languages, namely English, Arabic and Spanish, we then uncover large disparities in country representation, with the majority of data originating from a handful of countries. For each language, we also find that such countries are largely overrepresented in the HS datasets compared to their prominence both in the broader social media population and the general population of speakers. While the cross-geographic disparities in resources for certain languages had been discussed in past work \cite[e.g.,][]{arango-monnar-etal-2022-resources}, our work is the first to quantify such disparities and expose the lack of representativeness of existing resources.

\paragraph{Reasons for bias}
An important reason for the lack of representativeness of HS datasets comes from their primary data source, Twitter, which itself is a highly non-uniform sample of the broader social media population and the general population \cite{mislove2011understanding, lasri2023large}. In this regard, while our analysis exclusively focuses on Twitter, our findings are likely applicable beyond Twitter, as other data sources, such as Reddit, suffer from the same lack of representativeness.\footnote{\url{https://worldpopulationreview.com/country-rankings/reddit-users-by-country}} Beyond the data source, we observe that sampling decisions made by dataset creators are crucial in reducing representativeness. For instance, seed words are sometimes specific to certain countries, such as Chile for Spanish \cite{arango-monnar-etal-2022-resources}. 

\paragraph{Implications}

The primary implication of our work is the higher risk for less represented cultural contexts to face HS detection errors, due to several factors. First, HS often manifests in culturally specific forms, from its targets \cite{ousidhoum2021importance} to country-specific offensive words \cite{ghosh-etal-2021-detecting}. For instance, the Fulani ethnic group is an important target of online HS in Nigeria \cite{aliyu2022herdphobia, tonneau2024naijahate} whereas it is not in the US or the UK. The fact that such terms are likely to be less encountered during training may contribute to more false negatives and therefore less protection from HS in under-represented contexts \cite{dixon2018measuring}. Further, the same words could have different meanings across cultural contexts. For instance, \citet{castillo-lopez-etal-2023-analyzing} highlight the diverse connotations of the word ``fregar'' across Spanish-speaking regions, potentially carrying a misogynistic undertone in Spain but not in Ecuadorean Spanish. This discrepancy can lead to false positives and excessive moderation in under-represented contexts resulting from the application of cultural norms from over-represented contexts to under-represented contexts.

Moreover, this performance gap is compounded by a potential misalignment between the origins of data and annotators, resulting in a higher risk of annotation errors for less-represented countries in the annotation workforce. In this regard, we show that creators of English HS datasets seem less aware of this problem compared to Spanish or Arabic, as they consistently fail to provide information on the cultural contexts both the data and annotators originate from. A possible explanation for this difference is that contrary to English, dialects in some languages such as Arabic are not mutually intelligible (e.g., Moroccan and Syrian) rendering the match between data and annotator origin particularly relevant to ensure that the annotator understands the content they are supposed to annotate. Another possible explanation is the tendency to equate English with US-centric data as the majority of English tweets and researchers working on English HS originate from the United States, thereby overlooking the diversity of English speakers online.  This lack of information on data and annotator origins may hide a misalignment. For instance, 48\% of the crowdworkers employed by \citet{founta2018large} to annotate English tweets are from Venezuela. Lastly, we find that less developed countries tend to be under-represented in HS datasets, potentially reinforcing the marginalization of the same populations HS detection systems are built to protect. While this phenomenon has been documented within the US context for African Americans \cite{davidson-etal-2019-racial}, our findings suggest it can be extended globally.

\vspace*{-0.5em}

\paragraph{Recommendations}
Based on our results, we formulate three recommendations for the development of future HS datasets.

\begin{RecBox}{Recommendation 1}
Situate datasets in language and geography \end{RecBox}

\vspace*{-0.5em}

\noindent When possible, we argue that such a step is necessary to reduce cross-cultural errors in HS detection, especially for culturally-widespread languages such as English. This can be operationalized by using context-specific seed words for sampling or restricting the analysis to users with a specific location. It will allow practitioners to use data that corresponds to the cultural context they want to apply their models in. This additional information will also help better quantify the cultural bias in HS datasets and identify low-resource contexts that require more annotated data.

\begin{RecBox}{Recommendation 2}
Work with annotators that share the same origin as the data to annotate and specify their demographics 
\end{RecBox}

\noindent This second step will help further reduce detection errors, by ensuring that cultural nuances are well understood. Again, this is especially relevant for culturally-widespread languages and we acknowledge that this recommendation only holds in cases where the data's geographical origin is available or can be inferred. This is in line with prior work advocating for the inclusion of affected communities in determining what is hateful \cite{maronikolakis-etal-2022-listening} and also echoes the necessity of well-documented data statements \cite{bender-friedman-2018-data}.

\begin{RecBox}{Recommendation 3}
Ensure data availability while protecting user privacy 
\end{RecBox}

\noindent We find that a non-trivial amount of datasets cannot be retrieved. While it is crucial to protect the privacy of users on such a sensitive topic, ensuring data access is also crucial to maximize HS detection performance. In line with prior work \cite{assenmacher2023end}, we recommend to publicly release an anonymized version of the dataset and provide full data upon request, under conditions that protect users.

\section{Conclusion}

This work presented the first evaluation of cultural bias in HS datasets. We confirm past findings on the English-language bias of HS datasets, but also show that this bias has been steadily decreasing in the past few years. We also find evidence of geo-cultural bias for English, Arabic and Spanish, with HS datasets overrepresenting more developed countries and underrepresenting less developed countries. We finally uncover a relative lack of awareness of the possibility of such bias among English HS dataset creators, who systematically fail to provide information about data and annotator origin, hiding potential mismatches. Based on our results, we call for a more nuanced approach to HS detection that takes into account the specific cultural contexts in which speech occurs. We highlight that both language and geography are imperfect representations of culture on their own and discuss the importance of situating datasets using both features and resort to annotators sharing the same origin as the data to limit cross-cultural errors. Still, we are aware that what constitutes ``culture'' is debated \cite[e.g.,][]{kuper2000culture}, as are the rights of minority cultures vis-à-vis larger ones. We advocate for more inclusive representation of different cultures in resources like HS datasets, while recognizing the limitations of language and geography as cultural proxies.

\section*{Limitations}

\paragraph{Missing data}

An important limitation of our work is the sole focus on Twitter for the evaluation of geo-cultural bias. While we believe that our conclusions extend to other geo-culturally biased data sources of HS datasets (e.g., Reddit), we cannot empirically verify this claim. Further, we are only able to retrieve geographical information for a subset of all tweets and Twitter users. For instance, we cannot retrieve information for tweets with unavailable IDs, that were deleted or that do not have any geographical metadata. This data is likely not missing at random and thus represents a source of bias in our analysis. For instance, there may be a selection bias where users from some countries are more likely to share their location.

\paragraph{Location and geography do not equate culture} While we discuss the importance of using language and geography to define the origin of HS datasets, we are aware that both are imperfect proxies for culture. Diaspora communities illustrate this well: they often have a cultural mix from their origin and current countries. Also, users may provide incorrect location information.

\paragraph{Code-mixing} In our analysis, we only focus on single languages (e.g., English, Spanish). Yet, we are aware that code-mixing, that is the combined use of several languages, is prevalent in many English-speaking Majority World countries such as India and Nigeria. We are also aware that a few HS datasets exist for such contexts \cite[e.g.,][]{mathur-etal-2018-offend, tonneau2024naijahate} and encourage future work to include them in their analysis, in order to get a better estimate of cultural bias in HS datasets.

\section*{Ethical Considerations}

\paragraph{Data Privacy}

Owing to the sensitivity of the topic and to protect user privacy, we only provide aggregate results on user location. 

\section*{Acknowledgements}
We thank all the dataset authors who responded to our data requests and made this work possible. We also thank the reviewers for their feedback, which helped improve the paper.
PR is a member of the Data and Marketing Insights research unit of the Bocconi Institute for Data Science and Analysis, and is supported by a MUR FARE 2020 initiative under grant agreement Prot.\ R20YSMBZ8S (INDOMITA).

\bibliography{anthology,custom}
\bibliographystyle{acl_natbib}
\clearpage

\appendix

\section{Data Sources}
\label{sec:appendix_data_survey}

\subsection{Additional Descriptive Statistics}

We report the number of datasets by language and survey source in Table \ref{tab:survey_sources}. The main reason for dropping datasets from the analysis is that a lot of datasets do not focus specifically on hate speech but rather toxicity or offensiveness. The second main reason is the lack of availability of some datasets, as further detailed in \S \ref{sec:appendix_unavail_datasets}

\begin{table*}[htp]
\centering
\footnotesize
\resizebox{0.75\textwidth}{!}{%
\begin{tabular}{lrrrrrr}
\toprule
 \textbf{Language} & \makecell{HS Data \\ Catalogue} & \makecell{\citet{poletto2021resources}} & \makecell{Google Search} & \makecell{Total \\ found} &  \makecell{Total \\ kept}  \\
\midrule 
English    & 52 & 16 & 7  &  \textbf{75} &  \textbf{29}  \\
Arabic & 7 & 1 & 8 &  \textbf{16} &  \textbf{12} \\
Spanish   & 3 & 0 & 6 &   \textbf{9} &  \textbf{7}  \\
German   & 6 & 1 & 3 &   \textbf{10} &  \textbf{7}  \\
Turkish & 2 & 0 & 5 &   \textbf{7} &  \textbf{6} \\
French   & 3 & 1 & 4 &  \textbf{8} &  \textbf{6} \\
Portuguese   & 4 & 1 & 6 &  \textbf{11} &  \textbf{5} \\
Indonesian   & 3 & 0 & 4 &  \textbf{7} &  \textbf{3} \\
\bottomrule
\end{tabular}}
\caption{Number of available hate speech datasets by language and data source}
\label{tab:survey_sources}
\end{table*}

We also provide additional information in Table \ref{tab:post_count} on the total number of data points annotated for hate speech as well as the share of all data points by language. 

\begin{table}[htp]
\centering
\footnotesize
\resizebox{0.5\textwidth}{!}{%
\begin{tabular}{lrrr}
\toprule
 \textbf{Language} & \makecell{\# datapoints \\ in HS datasets} & \makecell{Share of \\ all HS datapoints} \\
\midrule 
English    & 623,272 & 41\%    \\
Arabic & 478,326 & 32\%   \\
Turkish & 151,921 & 10\%   \\
German   & 120,085 & 8\%   \\
Spanish   & 48,861 & 3\%  \\
Portuguese   & 46,914 & 3\%   \\
French   & 25,486 & 2\%   \\
Indonesian   & 14,904 & 1\%  \\
\bottomrule
\end{tabular}}
\caption{Number and share of datapoints by language for hate speech datasets}
\label{tab:post_count}
\end{table}

\subsection{Retained Hate Speech Datasets}
\label{sec:appendix_datasets}

We list below the retained datasets for each language, including six datasets under a ``Multilingual'' heading.
\paragraph{Arabic} 
\begin{enumerate}
    \item \textit{Are They Our Brothers? Analysis and Detection of Religious Hate Speech in the Arabic Twittersphere} \cite{albadi2018they}: 6,136 annotated Arabic tweets sampled using names of religious groups. Tweets are annotated as containing religious hate or not and for the hateful ones, which religious group is targeted. Religious hate is defined as ``speech that is insulting, offensive or hurtful and is intended to incite hate, discrimination, or violence against an individual or a group of people on the basis of religious beliefs or lack of any religious beliefs''. The annotators are CrowdFlower Arabic-speaking crowdworkers with an IP address in the Middle East. The inter-annotator agreement rate is 81\% for the first question and 55\% for the second question. 
    \item \textit{T-HSAB: A Tunisian Hate Speech and Abusive Dataset} \cite{haddad2019t}: 6,039 Tunisian Arabic social media posts sampled using hate-related keywords. The comments were annotated as either hateful, abusive or normal by three Tunisian native speakers with a higher education level. Hate comments are defined as instances that ``(a) contain an abusive language, (b) dedicate the offensive, insulting, aggressive speech towards a person or a specific group of people and (c) demean or dehumanize that person or that group of people based on their descriptive identity (race, gender, religion, disability, skin color, belief)''. The reported Krippendorff $\alpha$ is 0.75. 
    \item \textit{L-HSAB: A Levantine Twitter Dataset for Hate Speech and Abusive Language} \cite{mulki-etal-2019-l}: 5,846 Levantine tweets sampled using hate-related keywords. The comments were annotated as either hateful, abusive or normal by three Levantine native speakers with a higher education level. Hate comments are defined as instances that ``(a) contain an abusive language, (b) dedicate the offensive, insulting, aggressive speech towards a person or a specific group of people and (c) demean or dehumanize that person or that group of people based on their descriptive identity (race, gender, religion, disability, skin color, belief)''. The reported Krippendorff $\alpha$ is 0.765. 

    \item \textit{Hate and offensive speech detection on Arabic social media} \cite{alsafari2020hate}: 5,361 Gulf and Modern Standard Arabic tweets sampled through keyword-based, hashtag-based and profile-based approaches. The tweets are annotated in terms of hatefulness, aggressiveness, offensiveness, irony, stereotype and intensity. Hate speech is defined as ``possessing one or more of the following characteristics: 1. Insulting or defaming a specific group by using derogatory adjectives words or slurs.; 2. Defending or justifying hate crime.; 3. Promoting and encouraging hate.; 4. Advocating superiority of one group over the other.; 5. Threatening and inciting violence.; 6. Negative and disparaging stereotypes.;  7. Irony and jokes to humiliate and ridicule the target based on their protected characteristic.; 8. Special cases: a) Self-attacking, where the speaker attacks his own protected characteristic with hateful words. b) Re-posting or quoting hateful content''. The annotators are three Gulf native speakers with a high educational level. The Cohen $\kappa$ ranges from 0.77 to 0.9 across annotation levels. 
    \item \textit{Hate Speech Detection in Saudi Twittersphere: A Deep Learning Approach} \cite{alshaalan-al-khalifa-2020-hate}: 9,316 Saudi Arabic tweets sampled using keyword and hashtags. The tweets were annotated as hateful or not in batches by Figure Eight crowdworkers, Saudi annotators and three freelancers familiar with the Saudi dialect. Hate speech is defined as ``language that attack a person or a group based on some characteristic such as race, color, ethnicity, gender, religion, or other characteristic''. The inter-annotator agreement rate is not reported. 
    
    \item \textit{AraCOVID19-MFH: Arabic COVID-19 Multi-label Fake News \& Hate Speech Detection Dataset} \cite{ameur2021aracovid19}: 10,828 Arabic tweets sampled using keywords in the context of COVID-19. The tweets are annotated as hateful or not, whether it gives advice, whether it is news or an opinion, whether it contains blame or other negative speech and whether it is worth fact-checking. It is annotated by only one expert annotator. 
    \item \textit{Let-Mi: An Arabic Levantine Twitter Dataset for Misogynistic Language} \cite{mulki-ghanem-2021-mi} 6,550 Levantine Arabic tweets replying to popular female journalists during the October 17th 2019 in Lebanon. Tweets are annotated by three Levantine native speakers as non-misogynistic or as one of seven misogynistic categories (discredit, derailing, dominance, stereotyping and objectification, sexual harassment, threat of violence and damning). Unanimous agreement was found on 5,529 tweets, majority agreement on 1,021 tweets and conflicts on 53 tweets.
    \item \textit{Working Notes of the Workshop Arabic Misogyny Identification (ArMI-2021)} \cite{mulki2021working} 9,833 Arabic tweets for misogyny identification composed of Modern Standard Arabic and several Arabic dialects including Gulf, Egyptian and Levantine. The Levantine dataset corresponds to the Let-Mi dataset while the multi-dialectal tweets were collected using anti-women hashtags and scraping misogynists' timelines. The annotation scheme is both binary (misogynystic or not) and multi-class, following the annotation scheme of the Let-Mi dataset. The Krippendorff $\alpha$ is 0.94 for the binary task and 0.67 for the multi-class task.
    
    \item \textit{Overview of OSACT5 Shared Task on Arabic Offensive Language and Hate Speech Detection} \cite{mubarak-etal-2022-overview}: Arabic tweets sampled from \citet{mubarak2023emojis}. Each tweet was annotated by three Appen crowdworkers as 1) offensive or not and for offensive tweets 2) into fine-grained hate speech types. Hate speech is defined as ``offensive language targeting individuals or groups based on common characteristics such as Race (including also ethnicity and nationality), Religion (including belief), Ideology (ex: political or sport affiliation), Disability (including diseases), Social Class, and Gender''. Cohen's $\kappa$ value is 0.82. 
    \item \textit{Hate Speech Detection in the Arabic Language: Corpus Design, Construction and Evaluation} \cite{ahmad2023hate}: 403,688 Jordanian Arabic tweets sampled using language, keyword and location filters, focusing on users located in Jordan's main cities. The tweets were annotated by native Jordanian Arabic speakers as either positive, neutral, offensive but not hateful or hateful. Hate speech is defined as ``as a form of discourse that targets individuals or groups on the basis of race, religion, gender, sexual orientation, or other characteristics''. Fleiss' $\kappa$ is 0.6. 
\end{enumerate}

\paragraph{English} 
\begin{enumerate}
    \item \textit{Hateful Symbols or Hateful People? Predictive Features for Hate Speech Detection on Twitter} \cite{waseem-hovy-2016-hateful}: 16,907 annotated English tweets using a decision list to identify offensive content, focusing on oppression of minorities. Labels include ``Racism/Sexism/Neither''. The tweets were first annotated by the two authors and later refined by an external annotator. Inter-annotator agreement is $\kappa$=0.84.
    
    \item \textit{Are You a Racist or Am I Seeing Things? Annotator Influence on Hate Speech Detection on Twitter} \cite{waseem-2016-racist}: 6,909 annotated English tweets as an extension of \citet{waseem-hovy-2016-hateful} dataset, with an overlap of 2,876 tweets. Labels include ``Racism/Sexism/Neither/Both''. Annotators are recruited from CrowdFlower without a background selection. The inter-annotator agreement is $\kappa$=0.57.
    
    \item \textit{Automated Hate Speech Detection and the Problem of Offensive Language} \cite{davidson2017automated}: 24,802 annotated English tweets. Hate speech is defined as language that is used to expresses hatred towards a targeted group or is intended to be derogatory, to humiliate, or to insult the members of the group, with an emphasis on context. Labels include ``Hate speech/Offensive but not hate/Neither''. Annotators are recruited from CrowdFlower and the inter-annotator agreement is 0.92.
    
    \item \textit{When Does a Compliment Become Sexist? Analysis and Classification of Ambivalent Sexism Using Twitter Data} \cite{jha-mamidi-2017-compliment}: 7,205 annotated English tweets focusing on different types of \textit{sexist} content. Original labels include ``Benevolent sexism/Hostile sexism/Others''. ``Hostile sexism'' (N=3,378) and ``Others'' (N=11,559) tweets were extracted from \citet{waseem-hovy-2016-hateful}. ``Benevolent sexism'' content (N=7,205) was annotated by three experts with an interannotator agreement of 0.74. 
    
    \item \textit{Detecting Online Hate Speech Using Context Aware Models} \cite{gao-huang-2017-detecting}: 1,528 annotated comments of 678 users from the Fox News website. Hate speech is defined as language which explicitly or implicitly threatens or demeans a person or a group based upon a facet of their identity such as gender, ethnicity, or sexual orientation. Labels include ``Hateful/Non-hateful'', annotated by two native English speakers with an interannotator agreement of 0.98.
    
    \item \textit{Hate Speech Dataset from a White Supremacy Forum} \cite{de-gibert-etal-2018-hate}: 10,568 annotated sentences from posts and threads from Stormfront. Hate speech is defined as (a) deliberate attack (b) directed towards a specific group of people while (c) motivated by aspects of the group’s identity. Labels contain ``Hate/No hate/Relation/Skip''. ``Relation'' refers to a sentence that would be considered hateful when used together with other sentences. Three expert annotators achieved an agreement of 90.97\%.
    
    \item \textit{Peer to Peer Hate: Hate Speech Instigators and Their Targets} \cite{elsherief2018peer}: 27,330 annotated English tweets identifying hate content, as well as hate instigator and target. Hate speech definition was in line with content guidelines of Facebook and Twitter. Each tweet was annotated (a) hateful or not and (b) as containing a direct attack towards the mentioned account or not, by three Crowdflower annotators. Inter-annotator agreement is 92.8\% and 82.6\% for the two classifications respectively.

    \item \textit{Hate Lingo: A Target-based Linguistic Analysis of Hate Speech in Social Media} \cite{elsherief2018hate}: This dataset consists of 28,318 Twitter posts labeled as ``directed'' hate speech targeting specific individuals or entities, and 331 posts categorized as ``generalized'' hate speech directed towards broader groups with common protected characteristics like ethnicity or sexual orientation. Each tweet was annotated by at least three independent annotators from Crowdflower, with a Krippendorff’s $\alpha$ of 0.622.
    
    \item \textit{Large Scale Crowdsourcing and Characterization of Twitter Abusive Behavior} \cite{founta2018large}: 80,000 tweets annotated for various types of inappropriate speech. Initially classified into seven categories - offensive, abusive, hateful, aggressive, cyberbullying, spam, and normal - the final labels used were ``Normal/Spam/Abusive/Hateful''. Annotators were recruited from CrowdFlower with the largest group (48\%) from Venezuela. Agreement of annotators was grouped in three categories, with approximately 55.9\% of tweets receiving ``overwhelming agreement'' (at least 80\% of the annotators agree).
    
    \item \textit{Anatomy of Online Hate: Developing a Taxonomy and Machine Learning Models for Identifying and Classifying Hate in Online News Media} \cite{salminen2018anatomy}: 5,143 comments annotated for hateful content from YouTube and Facebook videos published by news media. One author performed open coding to develop a taxonomy of four types of hateful language - ``Accusations/Humiliation/Swearing/Promoting Violence'' - and nine target categories (e.g., religion, political issues). Then two other researchers coded a random sample, achieving an overall agreement score of 75.3\%.
    
    \item \textit{A Benchmark Dataset for Learning to Intervene in Online Hate Speech} \cite{qian-etal-2019-benchmark}: Two aggregated HS intervention datasets collected from Gab posts (N=21,747) and Reddit comments (N=7,641) respectively. Each conversation segment was annotated by three annotators who were recruited from Amazon Mechanical Turk (MTurk). The annotations include hate speech classification and suggested intervention responses.  
    
    \item \textit{Constructing interval variables via faceted Rasch measurement and multitask deep learning: a hate speech application} \cite{kennedy2020constructing}: 50,000 annotated social media comments from YouTube, Twitter, and Reddit written primarily in English. Annotations span eight categories from counterspeech to genocide. Annotators, recruited from MTurk, were evaluated using the (a) infit mean-squared statistic (0.37-1.9) to assess bias of favoring certain responses, and (b) the percentage of comments where the identity group of the hate target was flagged (no less than 20\%). 
    
    \item \textit{Detecting East Asian Prejudice on Social media} \cite{vidgen-etal-2020-detecting}: 40,000 English tweets aimed at detecting content targeting the East Asian community during Covid-19. Tweets were categorized into five primary groups: ``hostility/criticism/counterspeech/discussions of prejudice/unrelated''. 20,000 of these tweets were further annotated with secondary labels such as threatening language, interpersonal abuse, and dehumanization. Trained annotators specializing in hate speech performed the annotations. Each tweet was annotated by two annotators with a Fleiss' $\kappa$ of 0.54.
    
    \item \textit{HateXplain: A Benchmark Dataset for Explainable Hate Speech Detection} \cite{mathew2021hatexplain}: A total of 20,148 annotated posts sourced from Twitter (N=9,055) and Reddit (N=11,093). Data were annotated by three annotators from three different perspectives: the basic (``hate/offensive/normal''), the target community, and the rationales (specific post components considered hateful). Each tweet was annotated by three annotators recruited from MTurk with a Krippendorff’s $\alpha$ of 0.46.
    
    \item \textit{Learning from the Worst: Dynamically Generated Datasets to Improve Online Hate Detection} \cite{vidgen-etal-2021-learning}: This synthetic dataset contains 41,255 entries annotated for hate speech and non-hate speech. Specific types of hate identified include derogation, animosity, threatening language, support for hateful entities, and dehumanization, with targets of hate also noted. Annotation was performed on an open-source web platform with each case labeled by 3-5 trained annotators, primarily British (60\%), with expert oversights.
    
    \item \textit{``Call me sexist, but...'' : Revisiting Sexism Detection Using Psychological Scales and Adversarial Samples } \cite{samory2021call}: This re-annotated dataset comprises 4,078 entries from existing Twitter samples focused on sexism. Annotations cover overall sexism, four specific sexist content categories including behavioral expectations, stereotypes and comparisons, endorsements and denials of inequality, and rejection of feminism, plus three phrasing categories: ``uncivil and sexist/uncivil but not sexist/civil``. All annotators were U.S.-based MTurkers. Five annotators rate each entry and the majority agreement rates were 81\% for content, 98.8\% for phrasing, and 100\% for overall sexism.

    \item \textit{HateCheck: Functional Tests for Hate Speech Detection Models} \cite{rottger-etal-2021-hatecheck}: This synthetic dataset consists of 3,728 entries designed for hate speech detection, featuring 29 functionalities across 11 classes, such as profanity usage and pronoun reference. A team of ten trained annotators were recruited to ensure data quality, achieving a high inter-annotator agreement with a Fleiss’ $\kappa$ score of 0.93.
    
    \item \textit{An Expert Annotated Dataset for the Detection of Online Misogyny} \cite{guest-etal-2021-expert}: This dataset includes 6,383 Reddit posts and comments labeled for misogyny using a hierarchical taxonomy with four misogynistic categories (e.g., Pejoratives, Treatment, Derogation, Gendered Attacks) and three non-misogynistic categories (e.g., Counterspeech, Non-misogynistic Attacks, None). Secondary and third-level labels were also included. UK-based native English speakers annotated the dataset. Each data entry was annotated by 2-3 annotators. Inter-annotator agreement varied, with Fleiss’ $\kappa$ ranging from 0.145 to 0.559 for categories and 0.484 for the binary task (misogynistic/non-misognistic).
    
    \item \textit{Introducing CAD: the Contextual Abuse Dataset} \cite{vidgen-etal-2021-introducing}: This dataset features 25,000 annotated Reddit entries for classifying online abuse into six primary categories: ``Identity-directed/Person-directed/Affiliation-directed/Counter Speech/Non-hateful Slurs/Neutral'', along with subcategories. Annotations also noted whether contextual information was necessary and included corresponding rationales. Instead of crowdsourcing, trained institutional annotators were recruited. Inter-annotator agreement for the primary categories, measured by Fleiss’ $\kappa$, averaged 0.583.
    
    \item \textit{ETHOS: an Online Hate Speech Detection Dataset} \cite{mollas_ethos_2022}:
    Two datasets comprising 998 binary-labeled hateful comments and 433 messages with detailed labels were collected from YouTube (via Hatebusters) and Reddit. Annotations were conducted on the Figure-Eight platform, assessing whether comments contained hate speech, incited violence, or targeted specific groups. Further, comments were categorized based on hate speech related to gender, race, national origin, disability, religion, and sexual orientation. Almost each comment was annotated by five different annotators. Fleiss' $\kappa$ scores varied, reaching 0.814 for the binary variable and up to 0.977 for disability-related hate speech.
    
    \item \textit{Hatemoji: A Test Suite and Adversarially-Generated Dataset for Benchmarking and Detecting Emoji-based Hate} \cite{kirk-etal-2022-hatemoji}: The study presented two datasets examining hateful online emojis. The first dataset contains 3,930 hand-crafted test cases, annotated as hateful or non-hateful by three trained annotators, achieving a Randolph's $\kappa$ of 0.85. The annotators represented three nationalities—Argentinian, British, and Iraqi—with one being a native English speaker. The second dataset includes 5,912 entries annotated by a team of 11 (including one quality control annotator). Each entry was initially classified by three annotators, with hateful entries further categorized into four types and targets of hate. The annotator team included seven British, and one each from Jordanian, Irish, Polish, and Spanish backgrounds, with nine being native English speakers. Randolph’s $\kappa$ scores for three rounds ranged from 0.902 to 0.938.
    
    \item \textit{Introducing the Gab Hate Corpus: defining and applying hate-based rhetoric to social media posts at scale} \cite{kennedy2022introducing}: This dataset comprises 27,665 posts from Gab, annotated for hate speech using a hierarchical typology that distinguishes between high-level hate-based rhetoric, defined as ``Language that intends to — through rhetorical devices and contextual references — attack the dignity of a group of people, either through an incitement to violence, encouragement of the incitement to violence, or the incitement to hatred'', targeted populations (e.g., race or ethnicity), differentiation between mere vulgarity or aggression and hate speech, and between implicit and explicit rhetoric. Undergraduate research assistants based in the US were trained to annotate the data. Inter-annotator agreement was measured using Fleiss’s $\kappa$ and Prevalence-Adjusted, Bias-Adjusted $\kappa$. Agreement scores for top-level categories are human degradation (0.23, adjusted 0.67), calls for violence (0.28, adjusted 0.97), and vulgar/offensive content (0.30, adjusted 0.79). 
    
    \item \textit{Free speech or Free Hate Speech? Analyzing the Proliferation of Hate Speech in Parler} \cite{israeli-tsur-2022-free}: This dataset consists of 10,000 annotated posts from Parler, scored on a Likert scale from 1 (not hate) to 5 (extreme or explicit hate). A group of 112 student annotators achieved a satisfactory agreement level of 72\% and a Cohen’s $\kappa$ of 0.44.
    
    \item \textit{SemEval-2023 Task 10: Explainable Detection of Online Sexism} \cite{kirk-etal-2023-semeval}: This dataset includes 20,000 social media comments from Reddit and Gab to identify online sexism. Sexism was categorized on three levels: binary (sexist or not sexist), detailed sub-categories (threats, harm plans and incitement, derogation, animosity, and prejudiced discussion), and 11 specific manifestations. Each social media entry was reviewed by three trained annotators who all self-identified as women. The annotator team included seven British, as well as Swedish, Swiss, Italian, and Argentinian annotators, with eight being native English speakers. For cases lacking unanimous agreement in binary judgments, or less than two-thirds consensus in sub-categories and detailed manifestations, expert reviewers were consulted to provide final labels.
\end{enumerate}

\paragraph{French} 

\begin{enumerate}
    \item \textit{An Annotated Corpus for Sexism Detection in French Tweets} \cite{chiril-etal-2020-annotated}: 11,834 tweets for detecting sexism. Sexist content was defined as directed/descriptive/reported assertions to the addressee. Each tweet was annotated by five student annotators with an average Cohen’s $\kappa$ of 0.72 for sexist content/non sexist/no decision categories, and 0.71 for direct/descriptive/reporting/non sexist/no decision.
    
    \item \textit{CyberAgressionAdo-v1: a Dataset of Annotated Online Aggressions in French Collected through a Role-playing Game} \cite{ollagnier-etal-2022-cyberagressionado}: 19 multiparty chat conversations from a role-playing game for high-school students were collected and annotated to determine the presence of hate speech, type of verbal abuse, and humor. Hate speech was defined as content that mocks, insults, or discriminates based on characteristics like color, ethnicity, gender, sexual orientation, nationality, religion, or others. The dataset was fully annotated by one expert, with a second annotator reviewing four conversations. Inter-coder agreement reached Cohen's Kappa scores of 98.4\% for hate speech, 91.5\% for verbal abuse, and 96.3\% for humor.
    
    \item \textit{Detection of Racist Language in French Tweets} \cite{vanetik2022detection}: 2,856 annotated tweets for racist content detection. The dataset was annotated by two French native speakers with a $\kappa$ agreement of 0.66. In the case of disagreement, a third annotator assigned the final label. 
    
\end{enumerate}

\paragraph{German} 

\begin{enumerate}
    \item \textit{Detecting Offensive Statements Towards Foreigners in Social Media} \cite{bretschneider2017detecting}: Three datasets sourced from Facebook (with sample sizes of 2,649; 2,641; and 546) and focused on cyberhate and offensive language, particularly hostility towards foreigners. Offensive statements, their severity, and targets were annotated by two human experts. The intercoder agreement Cohen’s $\kappa$ yielded scores of 0.78, 0.68, and 0.73 for the respective datasets
    
    \item \textit{Measuring the Reliability of Hate Speech Annotations: The Case of the European Refugee Crisis} \cite{ross2017measuring}: 541 annotated original tweets containing only textual content, specifically to detect hate speech related to the refugee crisis. Each part was annotated by two annotators with a Krippendorff’s $\alpha$ of 0.38.
    
    \item \textit{RP-Mod \& RP-Crowd: Moderator-and Crowd-Annotated German News Comment Datasets} \cite{assenmacher2021texttt}: 85,000 annotated comments from a German newspaper \textit{Rheinische Post}. Comments were annotated for various types of hate speech including sexism, racism, threats, insults, and profane language, as well as for organizational content and advertisements. Annotations were conducted by crowdworkers from the Crowd Guru platform. Each comment was reviewed by five (close to) native German annotators, resulting in a Krippendorff’s $\alpha$ interannotator agreement score of 0.19.
    
    \item \textit{DeTox: A Comprehensive Dataset for German Offensive Language and Conversation Analysis} \cite{demus-etal-2022-comprehensive}: This dataset consists of 10,278 German annotated tweets, defined as hate speech if they ``attack or disparage persons or groups based on characteristics such as political attitudes, religious affiliation, or sexual identity'', and distinct from toxicity. Each comment was evaluated by three student annotators. Interannotator agreement, assessed using Gwet’s Agreement Coefficient, ranged from 0.75 to 0.95 across different categories.
    
    \item \textit{Improving Adversarial Data Collection by Supporting Annotators: Lessons from GAHD, a German Hate Speech Dataset} \cite{goldzycher2024improving}: This adversarial synthetic HS dataset includes approximately 10,966 examples. Hate speech was defined as abusive or discriminatory language targeting protected groups or individuals as members of such groups, with ``poor people'' also recognized as a protected category. All annotators are native or highly competent German speakers. The interannotator agreement across various rounds ranged from 0.83 to 0.99.
\end{enumerate}

\paragraph{Indonesian} 

\begin{enumerate}
    \item \textit{Hate speech detection in the Indonesian language: A dataset and preliminary study} \cite{alfina2017hate}: This dataset comprises 713 tweets related to the 2017 Jakarta Governor Election, annotated as hate speech or non-hate speech. Hate speech categories was defined as hatred of religion/ethnicity/race/gender. Each tweet was annotated by three student annotators, each from different religious, racial, and gender backgrounds. Tweets subject to disagreements were excluded, resulting in a 100\% interannotator agreement for the included tweets.

    \item \textit{Hate Speech Detection on Indonesian Instagram Comments using FastText Approach} \cite{pratiwi2018hate}:  The dataset consists of 572 annotated Indonesian Instagram comments, with 286 labeled as ``HS'' (presumably indicating hate speech) and 286 labeled as ``not HS'' (non-hate speech). The annotations were done manually by three Indonesian annotators from diverse age and gender backgrounds. Comments with disagreement among annotators were removed, ensuring 100\% inter-annotator agreement for the included samples.

    \item \textit{Multi-Label Hate Speech and Abusive Language Detection in Indonesian Twitter} \cite{ibrohim-budi-2019-multi}: 13,169 Indonesian tweets with 7,608 labeled as non-hate and 5,561 labeled as hate speech. The annotations cover abusive language, hate speech detection, identification of the target, category, and level of hate speech. The annotations were performed by crowdsourced native Indonesian annotators with diverse religious, racial/ethnic, and residential backgrounds. Each tweet was annotated by 3 annotators, and only tweets with 100\% inter-annotator agreement on the final label were included.
\end{enumerate}

\paragraph{Multilingual}
\begin{enumerate}
    \item \textit{SemEval-2019 Task 5: Multilingual Detection of Hate Speech Against Immigrants and Women in Twitter} \cite{basile-etal-2019-semeval}: The dataset contains 19,600 annotated tweets, with 13,000 in English and 6,600 in Spanish, focused on hate speech against immigrants and women. The annotations identify the presence of hate speech, the level of aggressiveness, and the targeted group. Three annotators labeled the data. For the English dataset, the reported average confidence scores (combining inter-rater agreement and reliability) are 0.83 for hate speech detection, 0.70 for identifying the target group, and 0.73 for aggressiveness level. For the Spanish dataset, the average confidence scores are 0.89, 0.47, and 0.47 respectively.
    
    \item \textit{CONAN - COunter NArratives through Nichesourcing: a Multilingual Dataset of Responses to Fight Online Hate Speech} \cite{chung-etal-2019-conan}: The dataset contains 4,078 pairs of hate speech and counter-narrative text, with 1,288 pairs in English, 1,719 in French, and 1,071 in Italian. The synthetic dataset was created by crowdsourcing to NGOs in the UK, France, and Italy. Two annotators per language independently annotated all the counter-narratives. The inter-annotator agreement, measured by Cohen's $\kappa$, is 0.92 across the three languages for annotating the hate speech sub-topic.
    
    \item \textit{Overview of the HASOC track at FIRE 2019: Hate Speech and Offensive Content Identification in Indo-European Languages} \cite{mandl2019overview}: The datasets contain annotated Twitter and Facebook data for hate speech detection in Hindi (N=4,665), German (N=3,819), and English (N=5,852). The labels include binary hate speech detection, types of hate speech, and the targeted group (for English and Hindi only). Several junior annotators were recruited, and the overlap percentages between annotators for hate speech detection on a subset annotated twice were 72\% for English, 83\% for Hindi, and 96\% for German.
    
    \item  \textit{Multilingual and Multi-Aspect Hate Speech Analysis} \cite{ousidhoum-etal-2019-multilingual}: The dataset comprises 13,014 tweets in Arabic (N=3,353), English (N=5,647), and French (N=4,014), labeled via crowdsourced annotators from MTurk using a multi-level scheme. The annotations capture directness, hostility level, target, group, and the annotator's feeling aroused by the tweet. Each tweet was annotated by five annotators and the interannotator agreement is measured using Krippendorff's $\alpha$ with 0.153 for English, 0.244 for French, and 0.202 for Arabic.
    
    \item \textit{Multilingual HateCheck: Functional Tests for Multilingual Hate Speech Detection Models} \cite{rottger-etal-2022-multilingual}: The dataset contains synthetic test cases for detecting hateful speech across ten languages: Arabic, Dutch, French, German, Hindi, Italian, Mandarin, Polish, Portuguese, and Spanish. It comprises 36,582 test cases, out of which 25,511 (69.7\%) are labeled as hateful, and 11,071 (30.2\%) as non-hateful. Hate speech was defined as abuse targeted at a protected group based on age, disability, gender identity, race, national or ethnic origin, religion, sex, or sexual orientation. Each test case was reviewed by three native-speaking annotators. Annotator agreement was measured by the portion of disagreement where at least 2 out of 3 annotators disagreed with the expert gold label, ranging from 0.73\% for Italian to 21.22\% for French.
    
    \item \textit{Large-Scale Hate Speech Detection with Cross-Domain Transfer} \cite{toraman-etal-2022-large}: 200,000 human-labeled tweets, covering both English (N=100,000) and Turkish (N=100,000) languages. Hate speech was defined including not only hateful behavior but also frequently observed domains based on target groups (religion, gender, race, politics, and sports). The labels include ``hate speech/offensive/normal''. Each tweet was annotated by five student annotators. The inter-annotator agreement, measured by Krippendorff's $\alpha$ coefficient, is 0.395 for the English data and 0.417 for the Turkish data.
\end{enumerate}

\paragraph{Portuguese} 

\begin{enumerate}
    \item \textit{A Hierarchically-Labeled Portuguese Hate Speech Dataset} \cite{fortuna-etal-2019-hierarchically}: 5,668 Portuguese tweets sampled using hate-related keywords and profiles. The annotators are Portuguese native speakers who are Information Science students. Each tweet is annotated by three students as hateful or not, and if hateful, the type of hate speech is also annotated (e.g., sexism). Hate speech is defined as ``language that attacks or diminishes, that incites violence or hate against groups, based on specific characteristics such as physical appearance, religion, descent, national or ethnic origin, sexual orientation, gender identity or other, and it can occur with different linguistic styles, even in subtle forms or when humour is used''. Fleiss' $\kappa$ is 0.17. 
    
    \item \textit{Toxic Language Dataset for Brazilian Portuguese (ToLD-Br)} \cite{leite-etal-2020-toxic}: 20,818 Brazilian Portuguese tweets sampled using keywords, hashtags as well certain user profiles (e.g., Bolsonaro). Each tweet was annotated by three Brazilian university students as either LGBTQ+phobia, obscene, insult, racism, misogyny, xenophobia or neutral. The average Krippendorff's $\alpha$ is 0.55. 
    
    \item \textit{HateBR: A Large Expert Annotated Corpus of Brazilian Instagram Comments for Offensive Language and Hate Speech Detection} \cite{vargas-etal-2022-hatebr}: 7,000 Brazilian Instagram posts commenting content from major Brazilian politicians. Each comment was annotated by three annotators in three steps: 1) offensive or not and 2) intensity of offensiveness and 3) hate speech type. Following \citet{fortuna-etal-2019-hierarchically}, hate speech is defined as ``a kind of language that attacks or diminishes, that incites violence or hate against groups, based on specific characteristics such as physical appearance, religion, or others, and it may occur with different linguistic styles, even in subtle forms or when humor is used. Therefore, hate speech is a type of language used against groups target of discrimination (e.g., sexism, racism, homophobia).'' The annotators are Brazilians with a high education level. The average Cohen's $\kappa$ is 0.75 for offensiveness and 0.47 for intensity of offensiveness. 
    
    \item \textit{TuPy-E: detecting hate speech in Brazilian Portuguese social media with a novel dataset and comprehensive analysis of models} \cite{oliveira2023tupy}: 9,367 Brazilian Portuguese tweets sampled using hate-related keywords and random sampling. Each tweet was annotated by three individuals in two steps: 1) as aggressive or not, 2) if aggressive, assign to one hate speech category among ageism, aporophobia, body shame, capacitism, LGBTphobia, political, racism, religious intolerance, misogyny and xenophobia. Hate speech is defined as ``the use of language that attacks or degrades, incites violence, or promotes hatred against groups based on specific characteristics such as physical appearance, religion, national or ethnic origin, sexual orientation''. Annotators are Brazilian with a high level of education. The agreement rate is not reported. 
\end{enumerate}

\paragraph{Spanish} 

\begin{enumerate}
    \item \textit{Detecting and Monitoring Hate Speech in Twitter} \cite{pereira2019detecting}: 6,000 annotated tweets from Spain selected using hate keywords. The tweets were annotated by four annotators (one public servant and three graduates) as hateful or not and a fifth annotation was sought in case of disagreements (Cohen $\kappa$: 0.588). Hate speech is defined as ``a kind of speech that denigrates a person or multiple persons based on their membership to a group, usually defined by race, ethnicity, sexual orientation, gender identity, disability, religion, political affiliation, or views''.
    \item \textit{Detecting misogyny in Spanish tweets. An approach based on linguistics features and word embeddings} \cite{garcia2021detecting}: 7,682 Spanish tweets from both Spain and Latin America, annotated as either misogynous or not. The tweets were annotated by two annotators (Krippendorff $\alpha$: 0.69). 
    \item \textit{Multilingual Resources for Offensive Language Detection} \cite{arango-monnar-etal-2022-resources}: 9,834 annotated Chilean Spanish tweets sampled using hate-related Chilean keywords. Tweets were annotated by three native Chileans as either hate speech, insult, unintended or intentional profanity. Hate speech is defined as ``stereotypical language to offend minority groups such as women, immigrants, sexual or racial minorities''. The authors report an agreement rate higher than 90\% and a Krippendorff $\alpha$ higher than 0.7 for all labels.
    \item \textit{Analyzing Zero-Shot transfer Scenarios across Spanish variants for Hate Speech Detection} \cite{castillo-lopez-etal-2023-analyzing}: 4,000 Spanish tweets from both Spain and Latin America sampled using geolocation and hate-related keywords. The tweets were annotated by three Latin American native Spanish speakers as xenophobic, non-xenophobic or ambiguous (Cohen $\kappa$: 0.44, agreement rate: 88\%). A tweet is xenophobic if (i) ``The content of the tweet primarily targets immigrants as a group, or even a single individual, if they are considered to be a member of that group (and NOT because of their individual characteristics)'' and (ii) ``The content of the tweet propagates, incites, promotes, or justifies hatred or violence towards the target or a message that aims to dehumanize, hurt or intimidate the target''.
    \item \textit{HOMO-MEX: A Mexican Spanish Annotated Corpus for LGBT+phobia Detection on Twitter} \cite{vasquez-etal-2023-homo}: 11,000 Mexican tweets sampled using nouns indicative of the LGBTQ+ community. The annotators were composed of 11 Mexican and 1 Colombian individuals. Each tweet were annotated by four annotators as either ``LGBTQ+phobic'', ``not LGBTQ+phobic'' or ``irrelevant to the LGBTQ+ community'' (Cohen $\kappa$: 0.43). If annotated as LGBTQ+phobic, the tweets were further annotated by type of LGBTQ+phobia.
\end{enumerate}

\paragraph{Turkish} 

\begin{enumerate}
    \item \textit{Hate Speech Detection with Machine Learning on Turkish Tweets} \cite{mayda2021turkcce}: 1,000 annotated Turkish tweets, sampled using names of target groups. Labels include \textit{hate speech, offensive expression, none of the two}. Annotated by two evaluators and disagreements are annotated by a third annotator (agreement rate of 83.4\%). 
    
    \item \textit{Hate Speech Dataset from Turkish Tweets} \cite{mayda2021hate}: 10,224 annotated Turkish tweets, sampled using name of target groups (e.g., jews). Labels include hate speech, offensive speech, or neutral. The tweets classified as hate were further annotated into subclasses, including ethnic, religious, sexist, and political tags. Two annotators labeled tweets separately, reaching a 92.5\% agreement rate, later increased to 98.4\% after discussion. A third evaluator resolved remaining disagreements.
    
    \item \textit{A Turkish Hate Speech Dataset and Detection System} \cite{beyhan-etal-2022-turkish}: This work contributes two hate speech datasets: the Istanbul Convention Dataset and the Refugee dataset. Hate speech is defined as ``language that is used to express hatred towards a targeted group or is intended to be derogatory, to humiliate, or to insult the members of the group''. The annotation scheme has four parts: (1) whether the tweet has no, weak or strong offensive language, (2) stance towards the Istanbul Convention or Refugees (pro, against or neutral), (3) target group and (4) hate speech type (e.g., insult, exclusion). The Istanbul Convention Dataset is composed of 1,206 tweets selected using hashtags and keywords. It was annotated by three senior undergraduate students (Krippendorff $\alpha$: 0.84 for binary task and 0.82 for multi-class task). The Refugee Dataset is composed of 1,278 tweets selected using immigrant-related keywords. Part of it was annotated by the undergraduate students and another part was annotated by employees of the Hrant Dink Foundation. 
    
    \item \textit{Homophobic and Hate Speech Detection Using Multilingual-BERT Model} \cite{karayiugit2022homophobic}: 31,290 Turkish Instagram comments sampled from accounts often posting homophobic and more generally hateful comments. The comments are annotated as either homophobic, hateful or neutral. The posts were annotated by two researchers.
    
    \item \textit{SIU2023-NST - Hate Speech Detection Contest} \cite{arin2023siu2023}: Shared task contributing two Turkish hate speech datasets: 2,240 tweets on the Israel-Palestine conflict annotated by hate speech type, as well as how severe hateful cases are; 4,683 tweets on refugees annotated as hate speech or not, as well as how severe hateful cases are.
\end{enumerate}

\subsection{Unavailable Datasets}
\label{sec:appendix_unavail_datasets}

We were not able to retrieve 5 English \cite{nobata2016abusive,fersini2018overview, rezvan2018quality,sarkar2021chess, vidgen2020detecting}, 3 Indonesian \cite{aulia2019hate,pratiwi2019hate,asti2021multi}, 3 Portuguese \cite{maronikolakis-etal-2022-listening, carvalho-etal-2022-hate, carvalho2023expression}, 1 Spanish \cite{fersini2018overview} and 1 German \cite{maronikolakis-etal-2022-listening} datasets.

\subsection{Official Statistics}
\label{sec:appendix_official_stats}
For English, we use data on the number of speakers as a first or second language\footnote{\url{https://en.wikipedia.org/wiki/List_of_countries_by_English-speaking_population}}. In the absence of such detailed data for other languages, we use data on the number of native speakers by country for Spanish\footnote{\url{https://cvc.cervantes.es/lengua/espanol_lengua_viva/pdf/espanol_lengua_viva_2022.pdf}} and Arabic\footnote{\url{https://www.worlddata.info/languages/arabic.php}}.

\section{Geocoding Evaluation}
\label{appendix_geocoding_eval}

We provide the full results of the geocoding evaluation in Table \ref{tab:geocoding_evaluation}.


\begin{table*}
\centering
\resizebox{0.9\textwidth}{!}{%
\begin{tabular}{lccc}
\toprule
 & English & Arabic & Spanish \\
\midrule
Share of geocoded user locations & 59\% & 71\% & 66\% \\
Share of correct geocoding & 92\% & 94\% & 96\% \\
Share of non-geocoded user locations that could have been geocoded from the provided information & 14\% & 12\% & 16\% \\
\bottomrule
\end{tabular}%
}
\caption{Geocoding evaluation}
\label{tab:geocoding_evaluation}
\end{table*}

\section{Comparison with Twitter Day}
\label{comparison_twitter_day}

\paragraph{Post-level} 

\begin{figure*}[t]
    \raggedleft
    \includegraphics[width=1\textwidth]{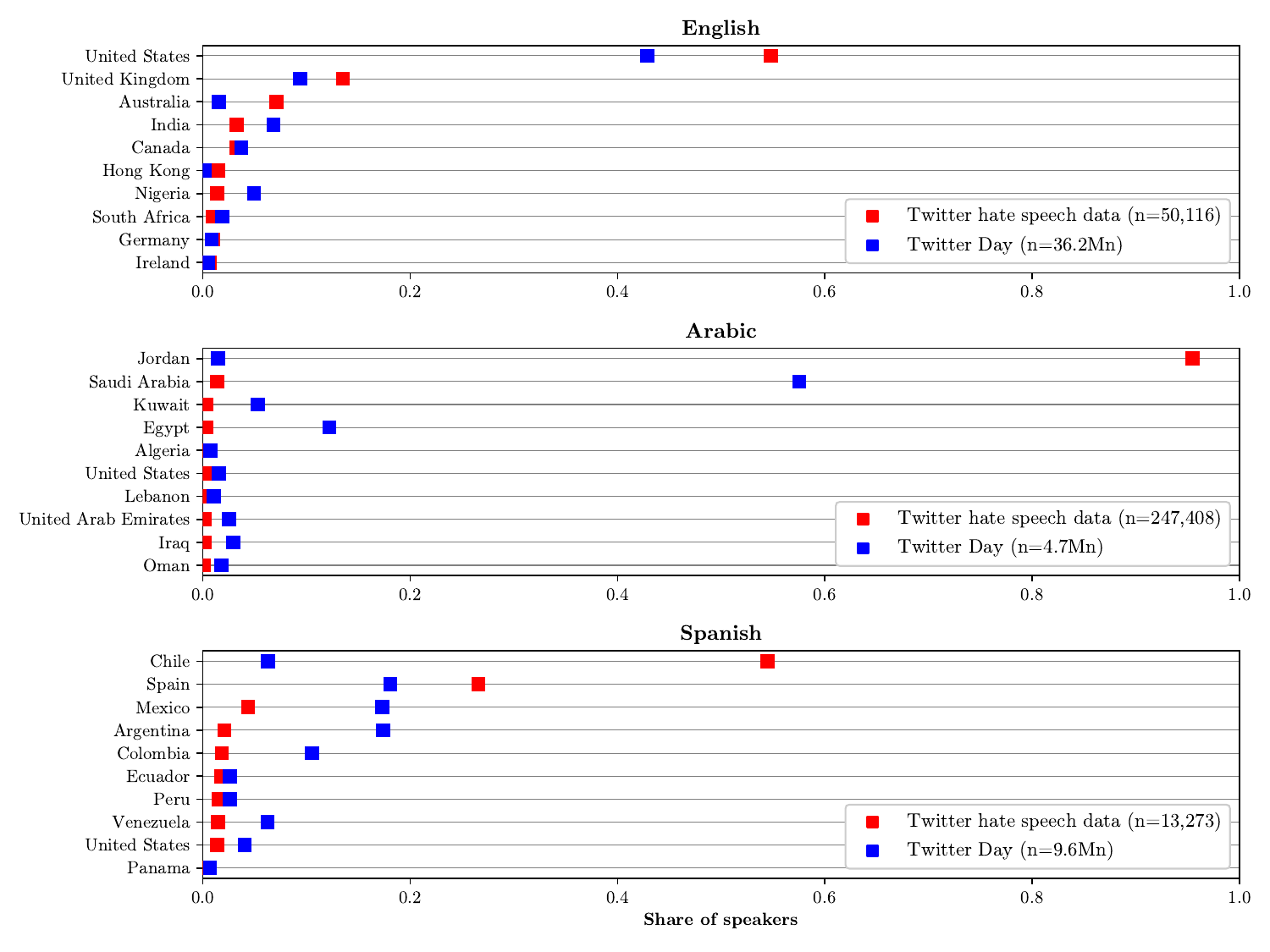} 
    \caption{Share of posts by country location in two reference populations: posts in the Twitter public hate speech datasets (Twitter hate speech data) and all Twitter posts, using the Twitter Day dataset as a proxy (Twitter Day) }
    \label{fig:masterplot_post_cross_country_twitter_day}
\end{figure*}

We provide a comparison between the country shares for posts in the Twitter hate speech data and the Twitter Day dataset in Figure \ref{fig:masterplot_post_cross_country_twitter_day}.

\begin{figure*}[t]
    \raggedleft
    \includegraphics[width=1\textwidth]{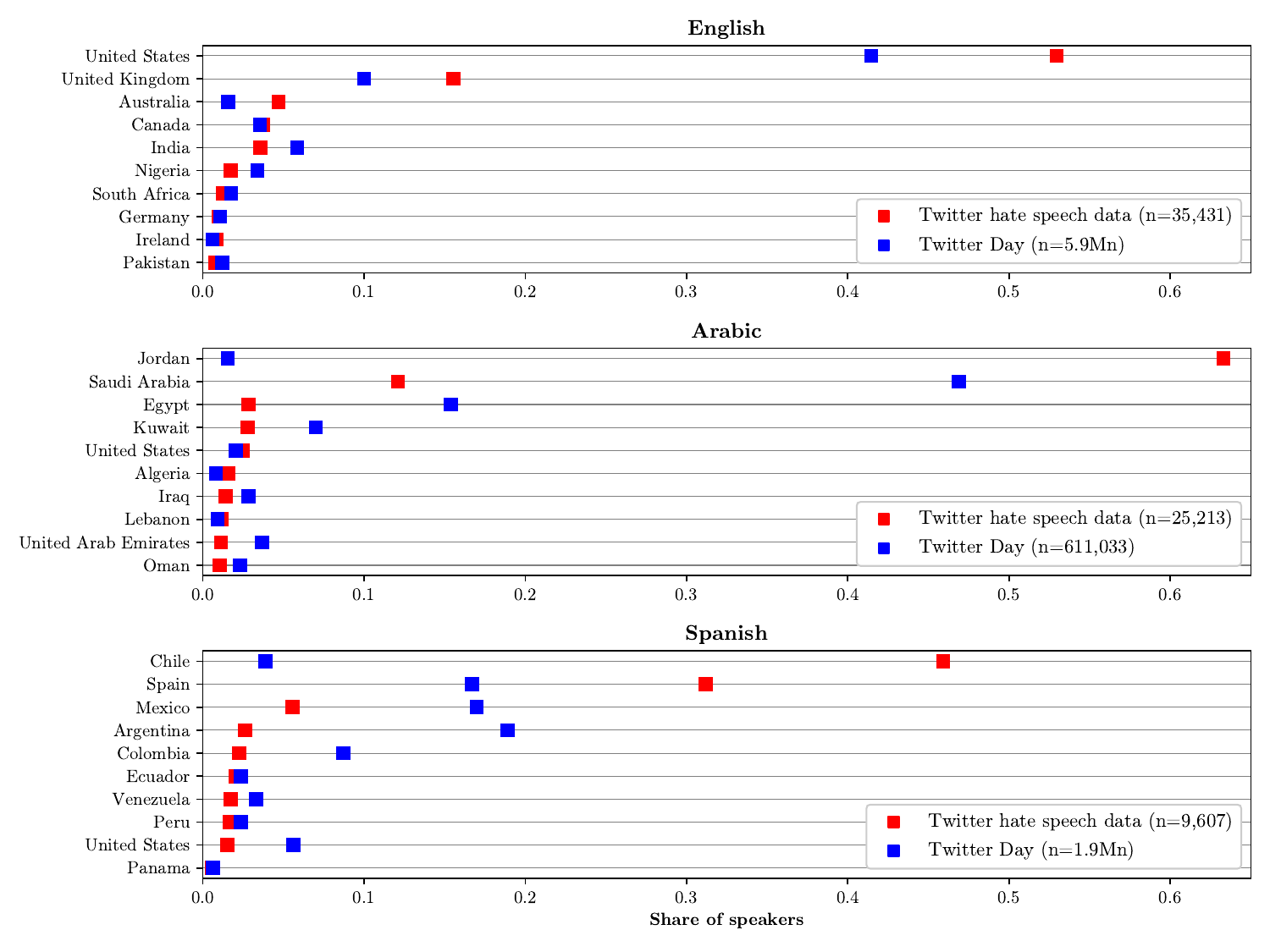} 
    \caption{Share of speakers by country location in two reference populations: Twitter users who authored the posts in the Twitter public hate speech datasets (Twitter hate speech data) and Twitter user population, using the Twitter Day data as a proxy (Twitter Day) }
    \label{fig:masterplot_cross_country_twitter_day}
\end{figure*}

\paragraph{User-level} We provide a comparison between the country shares for users in the Twitter hate speech data and in the Twitter Day datasets across languages (Figure \ref{fig:masterplot_cross_country_twitter_day}). 

\end{document}